\documentclass{article}
\usepackage{ijcai20}

\usepackage{times}
\usepackage{soul}
\usepackage{url}
\usepackage[hidelinks]{hyperref}
\usepackage[utf8]{inputenc}
\usepackage[small]{caption}
\usepackage{graphicx}
\usepackage{amsmath, amssymb}
\usepackage{amsthm}
\usepackage{booktabs}
\usepackage{algorithm}
\usepackage{algorithmic}
\usepackage{todonotes}
\usepackage{multirow}
\usepackage[linesnumbered,ruled,vlined,commentsnumbered, algo2e]{algorithm2e}
\newtheorem{definition}{Definition}
\newtheorem{theorem}{Theorem}

\newcommand{\ptitle}[1]{\vspace{1mm}\noindent{\bf #1.}}
\newcommand{\ptitlenoskip}[1]{\noindent{\bf #1.}}
\newcommand{\acronym}{ALPINE}

\usepackage{amsmath,amsfonts,bm}

\def\eqref#1{equation~\ref{#1}}

\def\1{\bm{1}}

\def\vx{{\bm{x}}}

\def\mA{{\bm{A}}}

\def\mC{{\bm{C}}}

\def\mX{{\bm{X}}}

\DeclareMathAlphabet{\mathsfit}{\encodingdefault}{\sfdefault}{m}{sl}
\SetMathAlphabet{\mathsfit}{bold}{\encodingdefault}{\sfdefault}{bx}{n}

\def\gG{{\mathcal{G}}}

\def\sR{{\mathbb{R}}}

\title{ALPINE: Active Link Prediction using Network Embedding}

\author{
Xi Chen
\and
Bo Kang\and
Jefrey Lijffijt\And
Tijl De Bie
\affiliations
Dept. of Electronics and Information
Systems, IDLab, Ghent University
\emails
\{xi.chen, bo.kang, jefrey.lijffijt, tijl.debie\}@ugent.be}

\begin{document}

\maketitle

\begin{abstract}
Many real-world problems can be formalized as predicting links in a partially observed network.
Examples include Facebook friendship suggestions, consumer-product recommendations,
and the identification of hidden interactions between actors in a crime network.
Several link prediction algorithms, notably those recently introduced using network embedding,
are capable of doing this by just relying on the observed part of the network.

Often, the link status of a node pair can be queried,
which can be used as additional information by the link prediction algorithm.
Unfortunately, such queries can be expensive or time-consuming, mandating the careful consideration of which node pairs to query.
In this paper we estimate the improvement in link prediction accuracy after querying any particular node pair,
to use in an active learning setup.

Specifically, we propose \acronym{} (Active Link Prediction usIng Network Embedding),
the first method to achieve this for link prediction based on network embedding.
To this end, we generalized the notion of $V$-optimality from experimental design to this setting,
as well as more basic active learning heuristics originally developed in standard classification settings.
Empirical results on real data show that \acronym{} is scalable, and boosts link prediction accuracy with far fewer queries.
\end{abstract}

\section{Introduction}\label{sec:introduction}
Applications of \emph{Link Prediction} (LP) in networks range from predicting social network friendships, consumer-product recommendations, citations in citation networks, to protein-protein interactions.
Such networks are usually only partially observed: node pairs are either connected (or \emph{linked}), disconnected (or \emph{unlinked}), or of \textit{unknown} status.
Indeed, obtaining network connections is usually resource-intensive (e.g., wet lab experiments or questionnaires), so that many of them remain unknown. 
Moreover, in many real-world networks new nodes are continuously added, with very limited information on their connectivity to the rest of the network.
In such \emph{Partially Observed Networks} (PONs), LP algorithms can be deployed to predict the missing link status information.
When no attribute or meta-data is available for the nodes---the situation we focus on in this paper---this must be done relying solely on structural information~\cite{kashima2009link}, i.e., the observed part of the PON.

In some cases, a budget is available for querying an oracle for the link status of a limited number of node pairs.
For example, wet lab experiments can reveal missing protein-protein interactions, or questionnaires can ask consumers to indicate whether they have seen particular movies before.
Unfortunately, such queries can be expensive,
while the link status of some node pairs is more informative than those of others---queries must thus be chosen wisely.
Given a finite budget, an active learning strategy, identifying and prioriti\-zing the most informative queries,
is thus required for optimal LP accuracy of the unobserved part in the PON.

For LP tasks, NE methods have become increasingly popular, owing to their high accuracy as well as versatility for other downstream tasks.
Thus, in this paper we develop \acronym{} (Active Link Prediction usIng Network Embedding), the first active learning method for NE-based LP in PONs.
For concreteness, we derived \acronym{} for Conditional Network Embedding~\cite{kang2018conditional} (CNE),
which achieves the state-of-the-art LP accuracy \cite{kang2018conditional}. 
Moreover, as opposed to other popular NE methods (including all those based on random walks),
CNE can distinguish \emph{disconnected} node pairs from those with \emph{unknown} status.
Additionally, its objective function is easy to express analytically,
which allows principled mathematical derivations for our active learning query strategies.
Yet, it will be clear that \acronym{} can be derived also for a wide range of other NE-based LP methods.

Given a PON,
\acronym{} must thus quantify the usefulness of querying a node pair with unknown status.
We introduce different strategies for doing this.
First, generalizing the notion of V-optimality from experimental design and exploiting the notion of Fisher information,
we derive a principled measure that quantifies the reduction in variance on the link probabilities for node pairs with unknown status.
Second, we propose several other heuristic strategies similar to those used in the standard active learning literature.

\begin{figure}
\includegraphics[width=\columnwidth]{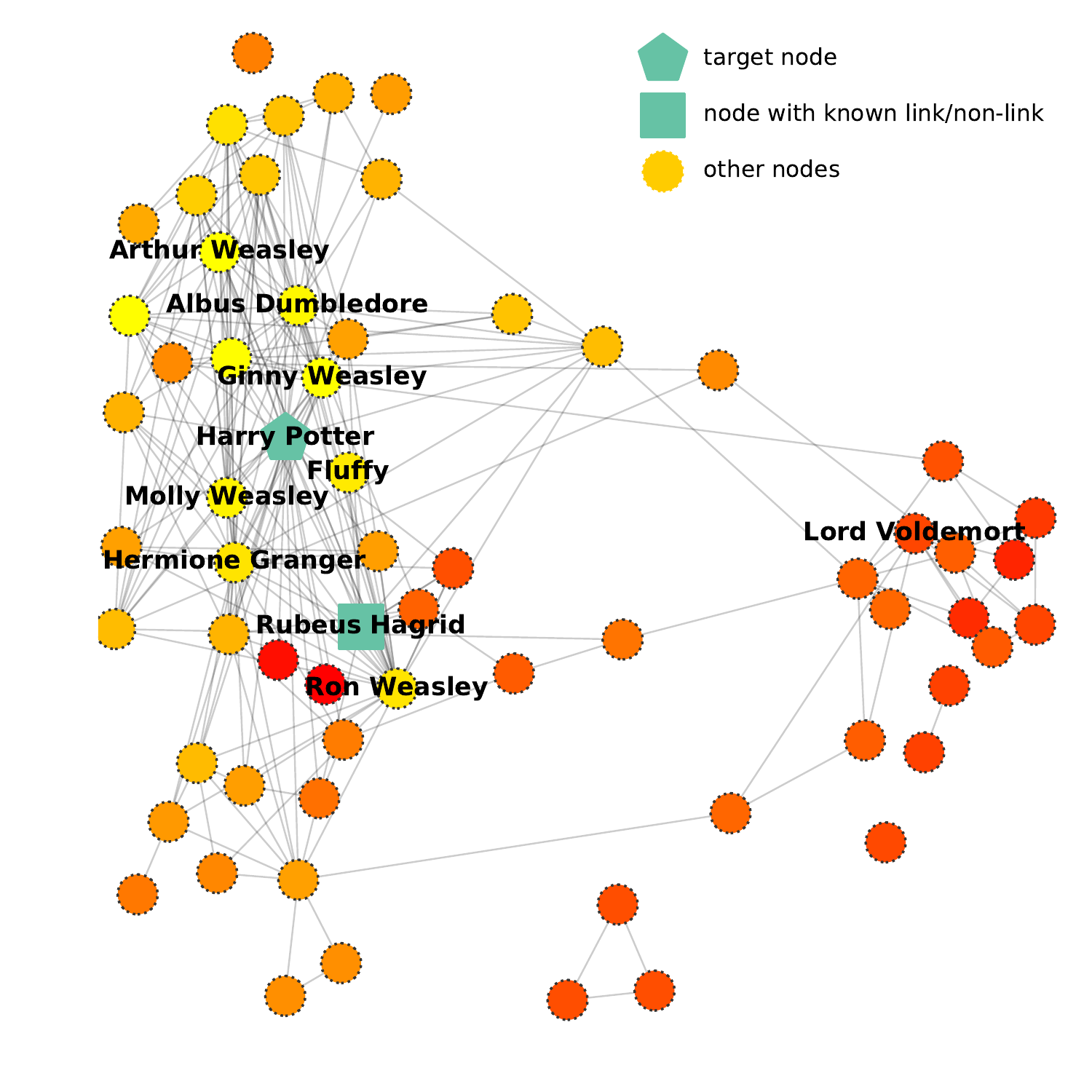}
\caption{Visualization of the Harry Potter network 
.\label{fig:potter}}
\end{figure}

\ptitlenoskip{Example} 
We illustrate the idea behind \acronym{} (with V-optimality query strategy) on the Harry Potter network \cite{Evans2014} of 65 Harry Potter characters and 221 ally connections amongst them (see Fig.~\ref{fig:potter}). 
We take node `Harry Potter' (pentagon node) and assume its connectivity is not observed except the connection to `Rubeus Hagrid' (square node). All other connections are assumed observed.
\acronym{} then scores the informativeness of the link status of `Harry Potter' to the other characters.
The round nodes are colored according to the scores: the yellower, the higher. 
\acronym{} selects the Weasley family (nodes with brightest yellow) as the top ones. 
Since the Weasley family is well connected with `Rubeus Hagrid' and it is also connected to many other characters,
it makes intuitive sense that the link status between them and `Harry Potter' is indeed very informative for predicting the other unknown connections for `Harry Potter'.

To the best of our knowledge, NE-based active LP has not been studied before. 
Indeed, while ample previous work on active learning for graphs exists,
this mainly focuses on node classification
~\cite{yang2014active,yang2016revisiting,bilgic2010active}.
The {\bf main contributions} of this paper are thus:
\begin{itemize}
\itemsep0em
  \item
    We highlight the importance of distinguishing the \emph{disconnected} versus \emph{unknown} status of node pairs, a rather obvious but often ignored fact
   (Sec.~\ref{sec:lp_pn}).
  \item
    We propose \acronym{}, an active learning approach for LP algorithms that are based on NE (Sec.~\ref{sec:alp}).
  \item
	 To identify the most informative node pairs to query, we generalize the notion of V-optimality to this setting.
	Moreover, we also propose a number of simpler heuristic query strategies inspired by active learning for standard learning settings.
  (Sec.~\ref{sec:query}.)
  \item
    Qualitative and quantitative evaluations
		show that \acronym{} with the V-optimality query strategy does indeed perform far better than random querying.
		Moreover, we show that two easy-to-compute heuristics achieve very similar performance,
		making them good alternatives on large networks.
(Sec.~\ref{sec:experiments}).
\end{itemize}

\section{Background}\label{sec:background}
Here we briefly survey active learning, (NE-based) link prediction, and some directly related work.

\subsection{Active Learning and Experimental Design}
\emph{Active learning} is a subfield of machine learning, which aims to exploit the situation where learning algorithms are allowed to actively choose the training data from which they learn.
It is particularly valuable in domains where training labels are scarce and expensive to acquire~\cite{brinker2003incorporating,cai2017active,settles2009active}.
The success of an active learning strategy depends on how much more effective its choice of training data is, when compared to randomly sampled training data.
Of particular interest to the current paper is the pool-based active learning,
where a pool of unlabeled data points is provided, and a subset from this pool can be selected for labeling.
In the context of the present paper, the unlabeled `data points' are the node pairs with unknown link status,
and an active learning strategy would aim to query the link status of those node pairs from this pool
that are most informative for a LP algorithm when used to predict the link status of the node pairs for which it is unknown.

Active learning is closely related to \emph{Optimal Experimental Design} (OED) in statistics~\cite{atkinson2007optimum},
which aims to design optimal `experiments' (i.e., the acquisition of training labels) with respect to a statistical criterion and within a certain cost budget. 
The objective of OED is usually to minimize a quantity related to the (co)variance matrix
of the estimated model parameters,
or of the predictions this model makes on the test data points.
In models estimated by the maximum likelihood principle,
a crucial quantity in OED is the \emph{Fisher Information}:
as the reciprocal of the estimator variance,
it allows quantifying the amount of information a training data point carries about the parameter.

While studied since long in statistics, the idea of variance minimization first shows up in the machine learning literature for regression~\cite{cohn1996active}, 
and later the Fisher Information was used to judge the asymptotic values of unlabeled data for classification~\cite{Zhang:Oles:00}. 
Yet, despite this related work in active learning, and the rich and mature statistical literature on OED for classification and regression problems,
to the best of our knowledge the concept of variance reduction has not yet been applied to LP in networks.

\subsection{Link Prediction and Network Embedding}
LP algorithms can be used in PONs to predict whether node pairs with unknown status are linked or not.
It has been widely applied in friendship recommendations, recommender systems, knowledge graph completion, and more.
While there are numerous conventional LP methods based on heuristic statistics~\cite{martinez2017survey},
recently proposed NE-based methods have been reported to outperform those~\cite{grover2016node2vec,kang2018conditional}.

Given a network $\mathcal{G}=(V,E)$ with nodes $V$ and links $E\subseteq\binom{V}{2}$,
the goal of NE is to find a mapping $f: V \to \sR^d$ from nodes to $d$-dimensional real vectors.
A NE is denoted as $\mX = (\vx_1, \vx_2,\ldots,\vx_n)' \in \sR^{n \times d}$, where $\mX^*$ denotes an \emph{optimal} embedding given the network $G$ with adjacency matrix $\mA$.
NE's can be used for a variety of downstream tasks, including visualization, node classification, and also LP.
When used for LP, a function $g: \sR^d \times \sR^d \to \sR$ evaluated on the vectors $\vx_i$ and $\vx_j$ might represent the probability (or other score) for a link to exist between $i$ and $j$. 
In practice, $g$ can be found by training a classifier (e.g., logistic regression) on a set of linked and unlinked node pairs, while in other cases it follows directly from the embedding model.
The method CNE used in this work is of the latter type, and aims to find an embedding that maximizes the probability of the network given the embedding:
\begin{equation*}
  P(\gG|\mX) = \prod_{\{i,j\} \in E} P(a_{ij}=1 | \mX) \cdot \prod_{\{k,l\}\notin E} P(a_{kl}=0 | \mX),
\end{equation*}
where $P(a_{ij} = 1|\mX)=g(\vx_i,\vx_j)$ for a suitably defined $g$.

Most NE methods treat node pairs with unknown status as unconnected.
For example, in methods based on skip gram with negative sampling (e.g, DeepWalk, node2vec), the random walks used to determine similarities between nodes use only known existing links, ignoring the unknown but potentially existing ones.
Those methods may therefore be suboptimal when applied to PONs,
and they cannot exploit new knowledge that a node pair with unknown status is not connected.
As we will show in Sec.~\ref{sec:method}, however, CNE can trivially be modified to distinguish these two situations---an important factor contributing to our decision to use CNE in this paper.

\subsection{Related work}
Our work sits at the intersection of three topics: active learning, link prediction, and network embedding.
There exists work on the \emph{pairs} of any two but not three.
To the best of our knowledge, \acronym{} is the first method for actively learning a NE for the purpose of LP.

\ptitle{NE-based LP} 
Many graph embedding methods have been proposed in the past years.
Based on neighborhood information, first order methods (e.g., CNE \cite{kang2018conditional}) and higher order methods (e.g., Deepwalk \cite{perozzi2014deepwalk}, node2vec \cite{grover2016node2vec}) have been designed to perform multiple tasks, such as LP and node classification.
Recently also Graph Convolutional Neural Networks (GCNNs) \cite{
kipf2016semi} were introduced, allowing nodes to recursively aggregate information from their neighbors. However, GCNNs mainly focus on node classification.

\ptitle{Active learning for NE} 
There are a few works on learning NE in an active manner recently, but they target node classification~\cite{cai2017active,chen2019activehne} instead of LP.

\ptitle{Active learning and link prediction} 
Work on active learning for graphs has focused on node and graph classification, as well as LP \cite{settles2009active,aggarwal2014active}.
The graph classification task considers data samples as graph objects, useful for drug discovery and subgraph mining~\cite{kong2011dual},
while the node classification task aims to label nodes in graphs~\cite{bilgic2010active,cesa2013active,guillory2009label}.
Other active learning research for LP considers different problem settings,
e.g., link classification for signed networks~\cite{cesa2012fast},
learning for graph edge flows~\cite{jia2019graph},
and training of the neural link predictor~\cite{ostapuk2019activelink}.
Probably the most strongly related method is HALLP~\cite{chen2014hallp},
which uses active learning for LP.
However, the method is heuristic,
it does not distinguish disconnected from unknown node pairs,
and it is based on a simple LP method that classifies node pairs according to a fixed representation of them in terms of engineered features,
rather than a learned NE.

\section{Method}\label{sec:method}
Section \ref{sec:lp_pn} formally defines PONs, and discusses how CNE is naturally suitable for embedding PONs.
Section \ref{sec:alp} describes \acronym{}, our NE-based active LP framework.
Section \ref{sec:query} shows how we generalize the notion of V-optimality from experimental design for \acronym{}, as well as more heuristic active learning query strategies.

\subsection{Link Prediction for PONs}\label{sec:lp_pn}
We formally define PONs as follows:
\begin{definition}
A Partially Observed Network (PON) is an undirected network $\gG = (V, E, U)$ where $V$ is a set of $n=|V|$ nodes,
$E \subseteq \binom{V}{2}$ and $U \subseteq \binom{V}{2}$ the sets of node pairs with connected and unknown status respectively,
where $E \cap U = \emptyset$.
$D\triangleq \binom{V}{2}\setminus (E\cup U)$ is the set of node pairs observed to be disconnected. Therefore, $K = E \cup D$ represents the observed part of the PON. 
\end{definition}
To represent three types of node pair status, the adjacency matrix $\mA$ of a PON has entries $a_{ij} \in \{0, 1, \text{null}\}$.

The task of LP in a PON is to predict the connectivity status of node pairs $(i,j)\in U$,
and this is based on the available information in $\gG$.
Remarkably, most NE methods (and LP methods more generally) do not treat disconnected node pairs differently from node pairs with unknown status.
This is inevitably true for methods based on random walks
(as a random walk cannot transition from node $i$ to $j$ if not known to be connected, regardless of whether known to be disconnected),
and true also for many other methods such as those based on matrix decompositions.
CNE, however, can be trivially modified to elegantly do so,
by maximizing the probability only for the observed part (i.e., $(i,j ) \in E\cup D$):
\begin{equation}\nonumber
  P(\gG|\mX) = \prod_{(i,j) \in E} P(a_{ij}=1 | \mX) \cdot \prod_{(k,l)\notin E\cup U} P(a_{kl}=0 | \mX).
\end{equation}

Furthermore, the link probability in CNE is formed analytically because the embedding is found by solving a Maximum Likelihood Estimation (MLE) problem: $\text{argmax}_\mX~ P(\gG | \mX)$.
Next, we will show how it allows us to quantify the informativeness of the node pairs with unknown status in \acronym{}.

\subsection{ALPINE}\label{sec:alp}
Here, we introduce \acronym{}, a pool-based active learning approach~\cite{settles2009active} for NE with LP as a downstream task.
We develop \acronym{} for CNE 
although we stress that our arguments can be applied in principle to any other NE method
of which the objective function can be expressed analytically.

\acronym{} works by finding an optimal NE for a given PON $\gG = (V, E, U)$,
selecting one or a few node pairs from $U$ to query,
updating the PON with the new knowledge (i.e. node pairs from $U$ found to be connected are moved to $E$,
those unconnected are removed from $U$),
and re-embedding the updated PON.
This process can be iterated until the budget is exhausted or until the model is sufficiently accurate.

We will introduce different strategies for selecting the node pairs to query,
relying on different \emph{utility functions} $u_{\mA, \mX}: V\times V \to \sR$ in \acronym{}
which quantify how useful knowing the connectivity status of a node pair is estimated to be
for the purpose of increasing the LP accuracy on the node pairs in $U$, when based on the updated NE (see Sec.~\ref{sec:query}).
Specifically, each query strategy will select the next query as:
\begin{align}\nonumber
	\underset{(i,j) \in U}{\text{argmax}}\ \ & u_{\mA, \mX}(i, j),
\end{align}
for an appropriate utility function $u_{\mA, \mX}$.
In practice, not just the single best node pair (argmax) is selected in each iteration,
but the $s$ best ones (further referred to as the `step size').

Thus, given a PON $\gG = (V, E,U)$, an NE model, a query strategy and associated utility function $u_{\mA,\mX}$,
a step size $s$, and a budget $B$ (number of node pairs in $U$ that can be queried), each iteration of \acronym{} works as follows.
We initialize the pool of node pairs with unknown status $U(0) = U$ and that of the known part $K(0)$ at step $it=0$ according to $\mA(0)$ of  $\gG (0) = \gG$ given.

\begin{enumerate}
    \item Compute $\mX^*(it)$ as an optimal embedding of the $\mathcal{G} (it)$;
    \item Find the best query $Q (it)\subseteq U (it)$ as the set of $|Q(it)|=\min(s, B)$ elements from $U(it)$ with largest values for the utility function $u_{\mA(it), \mX^*(it)}$;
    \item Query the oracle for the connectivity status of the node pairs in $Q(it)$, set $U(it+1)\leftarrow U(it)\setminus Q(it)$, and add each $(i,j)\in Q(it)$ revealed as connected to $E(it+1)$;
    \item Set $B\leftarrow B-|U(it)|$, and break if $B==0$.
\end{enumerate}
If desired, the LP accuracy based on the embedding can be monitored on a hold-out set during these iterations,
and one can stop early as soon as the accuracy meets a threshold.

\subsection{Query Strategies for \acronym{}}\label{sec:query}

Here we first derive a principled utility function based on the concept of V-optimality from OED.
The utility function measures the informativeness of the connectivity of a node pair
by identifying to what extent its knowledge is expected to minimize the variance of the predictions for the link status of node pairs in $U$. 
After that, we also introduce a range of other heuristic query strategies.

\ptitle{V-optimality and Variance Reduction}
V-optimality from OED aims to choose the training data points so as to minimize the variance of the predictions of the learned model on the test data points.
As the test set $U$ is finite and given in PONs, this aim naturally fits our problem setting.
Thus, with $g$ the link prediction function, and with $P^*_{ij}\triangleq g(\vx_i^*,\vx_j^*)=P(a_{ij}=1|\mX^*)$
the probability of a link between nodes $i$ and $j$ given the CNE embedding,
the utility function used in V-optimality is the reduction of $\sum_{(i,j) \in U} \text{Var} (P^*_{ij})$
achieved by querying a particular node pair from $U$.
The challenge to be addressed is thus the computation of the reduction in the variance terms $\text{Var} (P^*_{ij})$.
Omitting details, we outline how this can be done.

In \acronym{}, CNE finds the optimal embedding $\mX^*$ as the Maximum Likelihood Estimator (MLE)
given a PON with adjacency matrix $\mA$, i.e., $\mX^*$ maximizes $P(\gG|\mX)$ w.r.t. $\mX$.
The variance of an MLE can be quantified in terms of the Fisher Information~\cite{lehmann2006theory}.
More precisely, the Cramer-Rao bound~\cite{
rao1992information}
provides a lower bound on the variance of a MLE by the inverse of the Fisher Information:
$\text{Var}(\mX^*) \succeq \mathcal{I} (\mX^*) ^{-1}$.
Although the Fisher Information can often not be computed exactly (as it requires knowledge of the data distribution),
it can be effectively approximated by the \emph{observed} information matrix~\cite{efron1978assessing}.
For CNE, this observed information matrix \emph{for the MLE $\vx_i^*$} is given by (proof omitted for brevity):
\begin{align}\nonumber
\mathcal{I}(\vx_i^*) = \gamma^2 \sum_{(i,j) \notin U}  P_{ij}^*(1 - P_{ij}^*) (\vx_i^* - \vx_j^*) (\vx_i^* - \vx_j^*)^T,
\end{align}
where $\gamma$ is a CNE-parameter.
Thus, we can bound the covariance matrix $\mC_i$ of node $i$'s MLE embedding $\vx_i^*$ as $\mC_i \triangleq \text{Cov}(\vx_i^*)\succeq \mathcal{I}(\vx_i^*)^{-1}$.

Using a first-order analysis (details omitted) to decompose $\text{Var}(P_{ij}^*)$ into a
contribution from each end point as follows:
\begin{align}\label{eq:vardecomp}
\text{Var}(P_{ij}^*)&=\text{Var}_{\vx_i^*}(P_{ij}^*)+\text{Var}_{\vx_j^*}(P_{ij}^*),
\end{align}
and using the bound on $\text{Cov}(\vx_i^*)$ for all $i$,
allows bounding the variance on the estimated probabilities $\text{Var}(P_{ij}^*)$ by bounding the two terms in the decomposition as follows:
\begin{align}\label{eq:varbound}
\text{Var}_{\vx^*_i}(P_{ij}^*) \geq \left [ \gamma P_{ij}^*(1 - P_{ij}^*) \right ]^2 (\vx_i^* - \vx_j^*)^T \mC_i (\vx_i^* - \vx_j^*),
\end{align}
and similar for $\text{Var}_{\vx_j^*}(P_{ij}^*)$. Querying node pair $(i,j)\in U$ will reduce the covariance matrices $\mC_i$ and $\mC_j$,
as it creates additional information on their optimal values.
For example for $\vx_i^*$ (and similarly for $\vx_j^*$ leading to $\mC_j^i$), the new covariance assuming $(i,j)$ has known status, denoted $\mC_i^j$, is:
\begin{align}\label{eq:Cij}
\mC_i^j = \left [ \mC_i^{-1} + \gamma^2 P_{ij}^*(1 - P_{ij}^*) (\vx_i^* - \vx_j^*) (\vx_i^* - \vx_j^*)^T \right ]^{-1}.
\end{align}
Thus this leads to a reduction of the bounds on $\text{Var}_{\vx^*_i}(P_{ij}^*)$
and $\text{Var}_{\vx^*_j}(P_{ij}^*)$, and thus on $\text{Var}(P_{ij}^*)$ due to Eq.~(\ref{eq:vardecomp}).

Putting things together allows defining the V-optimality utility function,
and proves a theorem for computing it:
\begin{definition}
The V-optimality utility function $u_{\mA,\mX^*}$ evaluated at $(i,j)$ quantifies the reduction in the bound on the sum of the variances $\text{Var}(P_{kl}^*)$
(see Eqs.~(\ref{eq:vardecomp}) and~(\ref{eq:varbound})) of all $P_{kl}^*$ with $(k,l)\in U$, achieved by querying node pair $(i,j)\in U$.
\end{definition}
\begin{theorem}
The V-optimality utility function is given by:
\begin{align}\nonumber
u_{\mA,\mX^*}(i,j) &= \sum_{k:(i,k) \in U} u^{ik}(i,j) + \sum_{l:(j,l) \in U} u^{jl}(i,j),
\end{align}
where
\begin{align}\nonumber
u^{ik}(i,j) &= (\gamma P_{ik}^* (1-P_{ik}^*))^2  (\vx_i^* - \vx_k^*)^T (\mC_i-\mC_i^j) (\vx_i^* - \vx_k^*),\\\nonumber
u^{jl}(i,j) &= (\gamma P_{jl}^* (1-P_{jl}^*))^2  (\vx_j^* - \vx_l^*)^T (\mC_j-\mC_j^i) (\vx_j^* - \vx_l^*).
\end{align}
\end{theorem}
Applying the Sherman-Morrison formula
to Eq.~(\ref{eq:Cij}),
allows rewriting $u^{ik}(i,j)$ as:
\begin{align}\nonumber
\frac{\gamma^4 P_{ij}^* (1-P_{ij}^*)}{1+\gamma^2 P_{ij}^* (1-P_{ij}^*) d_{jj} (\vx_i^*)} \left [ P_{ik}^*(1-P_{ik}^*) \right ]^2 d_{kj} (\vx_i^*)^2.
\end{align}
where $d_{jj} (\vx_i^*) = (\vx_i^* - \vx_j^*)^T \mC_i (\vx_i^* - \vx_j^*) $ and $d_{kj} (\vx_i^*) = (\vx_i^* - \vx_k^*)^T \mC_i (\vx_i^* - \vx_j^*) $.
Thus, unsurprisingly, the variance reduction is always positive.

\subsubsection{Heuristic Query Strategy}
Besides V-optimality, we also propose 5 heuristic utility functions inspired by common active learning query strategies
(omitting subscripts from the utility function $u$ for brevity):
\begin{itemize}
\item \textbf{max-ent.}: $u(i,j) = -P_{ij}^*\text{log}P_{ij}^* - (1-P_{ij}^*)\text{log}(1-P_{ij}^*)$.
\item \textbf{max-prob.}: $u(i,j) = P_{ij}^*$.
\item \textbf{min-dis.}: $u(i,j) = - ||\vx_i^* - \vx_j^*||_2$.
\item \textbf{page-rank.}: $u(i,j) = \text{PR}_i  + \text{PR}_j $. 
\item \textbf{max-deg.}: $u(i,j) = \sum_{k:(i,k)\notin U} a_{ik} + \sum_{k:(j,k)\notin U} a_{jk}$.
\end{itemize}
The {\bf max-ent.} query strategy is a specific variant of the popular uncertainty sampling strategy in active learning, with the entropy as the uncertainty measure. 
The second and third strategies both tend to query node pairs that are linked with high probability. 
Indeed, this is true by definition for \textbf{max-prob.},
and approximately true for \textbf{min-dis.} as nearby nodes in the embedding are connected with higher probability.
The intuition behind these strategies is that links are often sparse in a network,
so that queries that result in the discovery of new links are more informative.
The last two are degree-related, and $\text{PR}_i $ in {\bf page-rank.} is the PageRank score of $i$ evaluated by treating node pairs with unknown status as disconnected.

\begin{figure*}[t]
\centering
\includegraphics[width=\textwidth]{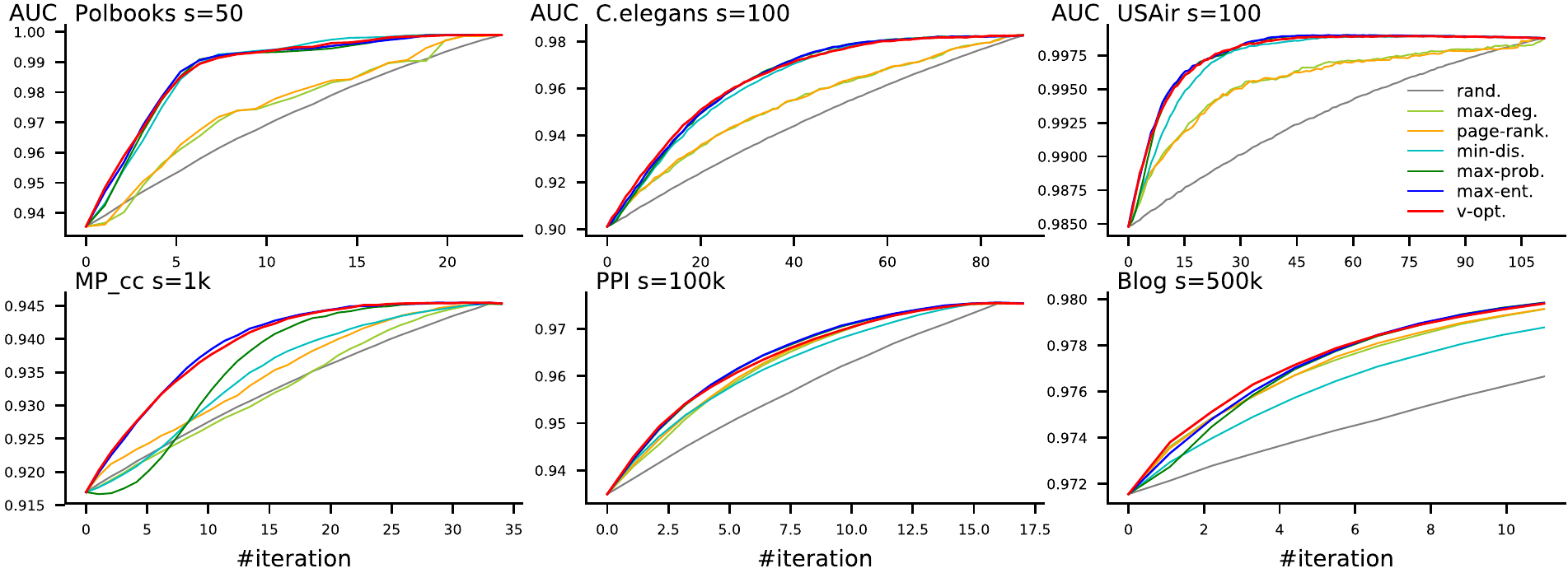}
\caption{Performance of the AL query strategies in PONs.\label{fig:quantitative_results}}
\end{figure*}

\begin{table}[tb]
\centering
\caption{\label{tab:budgeted} Percentage increase of AUC with a budget of $10\%$ of $|U|$.}
\resizebox{.48\textwidth}{!}{\begin{tabular}{cccc|ccc}
\hline
\multicolumn{1}{l}{\multirow{2}{*}{Query Strategy}} & \multicolumn{3}{c|}{Polbooks} & \multicolumn{3}{c}{C.elegans} \\ \cline{2-7}
\multicolumn{1}{l}{} & s=10 & s=50 & s=100 & s=10 & s=50 & s=1k \\ \hline
rand. & 0.87 & 0.81 & 0.88 & 1.05 & 1.03 & 1.04 \\
max-deg. & 0.57 & 0.70 & 0.70 & 1.80 & 1.54 & 1.32 \\
page-rank. & 1.01 & 0.88 & 0.84 & 1.71 & 1.45 & 1.35 \\
min-dis. & 1.97 & 1.91 & 2.09 & 2.06 & 1.65 & 1.36 \\
max-prob. & 1.94 & 2.13 & 2.17 & 2.14 & 1.57 & 1.29 \\
max-ent. & 2.17 & 2.28 & 2.29 & 2.25 & 1.73 & \textbf{1.37} \\
v-opt. & \textbf{2.35} & \textbf{2.34} & \textbf{2.33} & \textbf{2.44} & \textbf{1.88} & 1.36 \\
\hline
\multicolumn{1}{l}{\multirow{2}{*}{}} & \multicolumn{3}{c|}{USAir} & \multicolumn{3}{c}{MP\_cc} \\ \cline{2-7}
\multicolumn{1}{l}{} & s=100 & s=500 & s=1k & s=300 & s=1.5k & s=3k \\ \hline
rand. & 0.21 & 0.22 & 0.21 & 0.37 & 0.37 & 0.38 \\
max-deg. & 0.60 & 0.58 & 0.57 & 0.31 & 0.32 & 0.29 \\
page-rank. & 0.56 & 0.57 & 0.56 & 0.55 & 0.56 & 0.52 \\
min-dis. & 0.75 & 0.75 & 0.77 & 0.28 & 0.29 & 0.30 \\
max-prob. & 0.97 & 0.93 & 0.91 & 0.06 & 0.06 & 0.08 \\
max-ent. & \textbf{0.99} & \textbf{0.95} & \textbf{0.93} & 0.82 & 0.82 & \textbf{0.81} \\
v-opt. & 0.94 & 0.90 & 0.87 & \textbf{0.88} & \textbf{0.85} & 0.80 \\
\hline
\multicolumn{1}{l}{\multirow{2}{*}{}} & \multicolumn{3}{c|}{PPI} & \multicolumn{3}{c}{Blog} \\ \cline{2-7}
\multicolumn{1}{l}{} & s=10k & s=30k & s=50k & s=100k & s=300k & s=500k \\ \hline
rand. & 0.48 & 0.50 & 0.51 & 0.14 & 0.13 & 0.13 \\
max-deg. & 0.82 & 0.85 & 0.82 & 0.33 & 0.32 & 0.33 \\
page-rank. & 0.87 & 0.88 & 0.88 & 0.33 & 0.33 & 0.34 \\
min-dis. & 0.97 & 0.97 & 0.96 & 0.24 & 0.23 & 0.25 \\
max-prob. & 1.08 & 1.07 & 1.09 & 0.30 & 0.28 & 0.30 \\
max-ent. & 1.10 & 1.09 & 1.10 & 0.33 & 0.31 & 0.33 \\
v-opt. & \textbf{1.14} & \textbf{1.16} & \textbf{1.16} & \textbf{0.37} & \textbf{0.35} & \textbf{0.37} \\
\hline
\end{tabular}
}
\end{table}

\begin{table}[tb]
\centering
\caption{\label{tab:runtime} Runtime evaluation in seconds.}
\resizebox{.48\textwidth}{!}{\begin{tabular}{c|c|c|c|c|c|c|c}
 & rand. & max-deg. & page-rank. & min-dis. & max-prob. & max-ent. & v-opt. \\ \hline
Polbooks & $2\mathrm{e}{-4}$ & $2\mathrm{e}{-3}$ & $0.04$ & $7\mathrm{e}{-3}$ & $1\mathrm{e}{-3}$ & $9\mathrm{e}{-4}$ & $0.09$ \\
C.elegans & $2\mathrm{e}{-5}$ & $0.02$ & $0.10$ & $0.05$ & $0.01$ & $8\mathrm{e}{-3}$ & $0.75$ \\
USAir & $3\mathrm{e}{-5}$ & $0.02$ & $0.14$ & $0.07$ & $0.01$ & $0.01$ & $0.93$ \\
MP\_cc & $2\mathrm{e}{-5}$ & $0.06$ & $1.59$ & $0.19$ & $0.04$ & $0.03$ & $2.86$ \\
PPI & $3\mathrm{e}{-5}$ & $2.85$ & $3.86$ & $9.62$ & $2.53$ & $2.23$ & $255.61$ \\
Blog & $0.02$ & $25.57$ & $36.36$ & $74.04$ & $24.05$ & $7.71$ & $3\mathrm{e}{3}$ \\
\end{tabular}}
\end{table}

\section{Experiments}\label{sec:experiments}
We investigated the following questions:
{\bf Q1} Does the behaviour of \acronym{} make sense qualitatively?
{\bf Q2} Do the different query strategies perform well in predicting the node pairs with unknown status?
{\bf Q3} Do the query strategies scale to large networks?

\ptitlenoskip{Data}
\acronym{} is evaluated on 6 real datasets of varying sizes.
{\bf Polbooks} is a network consisting of 105 books about the US politics among which 441 connections indicate the co-purchasing relations between the book pairs~\cite{Adamic05Polbooks} .
{\bf C.elegans} is a neural network of C.elegans with 297 neurons and 2,148 synapses as their links~\cite{watts1998collective}.
{\bf USAir} is a network of 332 airports connected through 2,126 US Airlines~\cite{handcock2003statnet}.
{\bf MP\_cc} network is a twitter network we gathered in April 2019 for the Members of Parliament (MP) in the UK, which originally contains 650 nodes, and we only use its largest connected component of 567 nodes and 35531 friendship (i.e., mutual follow) connections.
{\bf PPI} is a protein-protein interaction network with 3,890 proteins and 76,584 interactions~\cite{breitkreutz2007biogrid}.
{\bf Blog} is a friendship network of 10312 bloggers from BlogCatalog, containing 333983 connections~\cite{Zafarani+Liu:2009}.

\subsection{Qualitative evaluation \label{sec:qualitative}}
As illustrated in Fig.~\ref{fig:potter}, we apply \acronym{} to the case when a new node is added to a network, and examine the behaviour of the V-optimality query strategy.
We take node 39 (`Harry Potter', pentagon node) as newly arrived who has only one initial known connection to node 22 (`Rubeus Hagrid', square node).
Thus, the node pairs involving Harry, except the one with Rubeus, all have unknown status.

In the first iteration, the V-optimality query strategy used in \acronym{} scores all the node pairs $(i,j) \in U$ and suggests to query (`Harry Potter', `Arthur Weasley') where `Arthur Weasley' (node with brightest yellow) is the father of the Weasley family.
The other members of Weasley family are also scored high.
This indicates that predicting the relationships of `Harry Potter' with other characters can be improved
by first querying possible connections with members of the Weasley family -- close allies of `Harry Potter' and with many connections to other characters.

From the second to the fifth iterations, the top ranked nodes are `Ginny Weasley', `Fred Weasley', `George Weasley', `Albus Dumbledore'.
These early suggestions sketch the relationships of `Harry Potter' to the entire network, thus allowing it to be well-embedded with just a few queries.

\subsection{Quantitative evaluation \label{sec:quantitative}}
We quantitatively evaluate \acronym{} with different query strategies for LP on node pairs with unknown status in a PON in an iterative manner.
All query strategies proposed in Sec.~\ref{sec:query} are used.
Additionally, a baseline query strategy which samples node pairs uniformly at random from $U$, is included for comparison.
In order to construct PON based on the benchmarks, information on $20\%$ of the node pairs (both links and non-links) is removed, which forms $U$.
Then we apply \acronym{} for varying budget $B$ and and a range of step sizes $s$.

Fig.~\ref{fig:quantitative_results} shows the LP accuracy over subsequent iterations for all datasets, where the first 5 are queried with no budget limitation (i.e., $B=|U|$) and {\bf Blog} with $B=5M$.
Hereby, the LP accuracy is quantified as the AUC of all node pairs in $U(0)$ for $it=0$ with respect to the ground truth.
Results are averaged over several $U(0)$s, and the PON with each $U(0)$ is initialized with different random embeddings ($10\times 10$ for the first four, $5\times 5$ for the fifth and $4 \times 2$ for the last).
Scores of the random strategy are further averaged over 5 runs.

All non-random strategies perform consistently and substantially better than the random query strategy, apparently clustering into two groups within which the accuracy is similar.
The 4 best strategies are all NE-based: {\bf v-opt.}, {\bf max-ent.}, {\bf max-prob.}, and {\bf min-dis}. With these strategies, \acronym{} boosts link prediction accuracy with far fewer queries.
One interesting finding is that {\bf max-prob.}, while different in spirit to uncertainty maximization, still performs similar to {\bf max-ent.}.
The reason could be that positive links are considered as more informative because real-world networks are usually sparse such that linked node pairs are more informative.
The second group consists of {\bf page-rank}, and {\bf max-deg.}, both strategies that do not require the NE.
Thus, the link status of high-degree nodes is more informative than that of the random ones, but as a strategy it is inferior to the NE-based ones.

In practice, however, active learning is particularly useful when the budget is small.
Thus, we investigated in greater detail the relative performance of the various query strategies for a small budget $B$, equal to $10\%$ of $|U|$.
Table~\ref{tab:budgeted} shows the increase in percentage points of the LP AUC compared to the AUC before active learning, and this for three different step sizes.
The V-optimality query strategy outperforms the others in most cases,
and is close second or third in a few other cases, although {\bf max-prob.} and {\bf max-ent.} are never much worse.
As a side result, the Table shows that the LP AUC is relatively insensitive to the step size.

We also evaluated \acronym{} for predicting the connectivity to a newly added node, using as few queries as possible, with similar conclusions
(details will be in an extended version).

\subsection{Scalability \label{sec:scalability}}
The runtime analysis (on a server with Intel Core i5 CPU 2.30GHz and 8GB RAM) of \acronym{} with different query strategies is shown in Table~\ref{tab:runtime}.
The embedding dimension is set to $8$, and the removed information from the original network is $20\%$ of the node pairs.
The results are averaged over $10$ random runs.
It shows that the computation time per iteration of the V-optimality strategy increases dramatically as the network size grows.
Given this, and their competitive performance in terms of LP accuracy, {\bf max-prob.} and {\bf max-ent.} are probably the query strategies of choice on larger problems.

\section{Conclusion}\label{sec:conclusion}
Link prediction is an important task in network analysis, tackled increasingly using network embeddings. 
It is particularly important in partially observed networks,
where finding out whether a pair of nodes is linked is time-consuming or costly,
such that for a large number of node pairs it is not known if they are connected or not.
 
We propose to make use of active learning in this setting, and introduce \acronym{}, a specific active learning approach for link prediction in such partially observed networks using network embedding.
We first derived a principled query strategy that generalizes the notion of V-optimality from optimal experimental design to the current setting,
identifying those node pairs which, if queried, will maximally reduce the variance on the link scores for the node pairs with unknown connectivity status.
Additionally, several heuristic active learning strategies are also proposed as computationally efficient alternatives.
Empirical evaluations show that \acronym{} with the V-optimality query strategy performs best overall, albeit at a relatively high computational cost,
while two intuitive heuristics achieve similar accuracies and scale to larger networks.
All query strategies outperform by a large margin the random query strategy.

As future work, we plan to further improve the scalability of \acronym{} by e.g., using incremental embeddings at each iteration.

\section*{Acknowledgments}
The research leading to these results has received funding from the European Research Council under the European Union's Seventh Framework Programme (FP7/2007-2013) / ERC Grant Agreement no. 615517, from the Flemish Government under the ``Onderzoeksprogramma Artifici{\"e}le Intelligentie (AI) Vlaanderen'' programme, from the FWO (project no. G091017N, G0F9816N, 3G042220), and from the European Union's Horizon 2020 research and innovation programme and the FWO under the Marie Sklodowska-Curie Grant Agreement no. 665501. We thank Ahmad Mel for helping collecting the MP network.

\bibliographystyle{named}
\bibliography{main}

\begin{thebibliography}{}

\bibitem[\protect\citeauthoryear{Adamic and Glance}{2005}]{Adamic05Polbooks}
Lada~A. Adamic and Natalie Glance.
\newblock The political blogosphere and the 2004 u.s. election: Divided they
  blog.
\newblock In {\em Proc. of the 3rd International Workshop on Link Discovery}.
  ACM, 2005.

\bibitem[\protect\citeauthoryear{Aggarwal \bgroup \em et al.\egroup
  }{2014}]{aggarwal2014active}
Charu~C Aggarwal, Xiangnan Kong, Quanquan Gu, Jiawei Han, and S~Yu Philip.
\newblock Active learning: A survey.
\newblock In {\em Data Classification}, pages 599--634. Chapman and Hall/CRC,
  2014.

\bibitem[\protect\citeauthoryear{Atkinson \bgroup \em et al.\egroup
  }{2007}]{atkinson2007optimum}
Anthony Atkinson, Alexander Donev, et~al.
\newblock {\em Optimum experimental designs, with SAS}, volume~34.
\newblock Oxford University Press, 2007.

\bibitem[\protect\citeauthoryear{Bilgic \bgroup \em et al.\egroup
  }{2010}]{bilgic2010active}
Mustafa Bilgic, Lilyana Mihalkova, and Lise Getoor.
\newblock Active learning for networked data.
\newblock In {\em Proc. of ICML}, pages 79--86, 2010.

\bibitem[\protect\citeauthoryear{Breitkreutz \bgroup \em et al.\egroup
  }{2007}]{breitkreutz2007biogrid}
Bobby-Joe Breitkreutz, Chris Stark, et~al.
\newblock The biogrid interaction database: 2008 update.
\newblock {\em Nucleic acids research}, 36:D637--D640, 2007.

\bibitem[\protect\citeauthoryear{Brinker}{2003}]{brinker2003incorporating}
Klaus Brinker.
\newblock Incorporating diversity in active learning with support vector
  machines.
\newblock In {\em Proc. of ICML}, pages 59--66, 2003.

\bibitem[\protect\citeauthoryear{Cai \bgroup \em et al.\egroup
  }{2017}]{cai2017active}
Hongyun Cai, Vincent~W Zheng, and Kevin Chen-Chuan Chang.
\newblock Active learning for graph embedding.
\newblock {\em arXiv preprint arXiv:1705.05085}, 2017.

\bibitem[\protect\citeauthoryear{Cesa-Bianchi \bgroup \em et al.\egroup
  }{2012}]{cesa2012fast}
Nicolo Cesa-Bianchi, Claudio Gentile, Fabio Vitale, and Giovanni Zappella.
\newblock A fast active learning algorithm for link classification.
\newblock In {\em Proc. of ICTCS}, 2012.

\bibitem[\protect\citeauthoryear{Cesa-Bianchi \bgroup \em et al.\egroup
  }{2013}]{cesa2013active}
Nicolo Cesa-Bianchi, Claudio Gentile, et~al.
\newblock Active learning on trees and graphs.
\newblock {\em arXiv preprint arXiv:1301.5112}, 2013.

\bibitem[\protect\citeauthoryear{Chen \bgroup \em et al.\egroup
  }{2014}]{chen2014hallp}
Ke-Jia Chen, Jingyu Han, and Yun Li.
\newblock Hallp: A hybrid active learning approach to link prediction task.
\newblock {\em JCP}, 9(3):551--556, 2014.

\bibitem[\protect\citeauthoryear{Chen \bgroup \em et al.\egroup
  }{2019}]{chen2019activehne}
Xia Chen, Guoxian Yu, et~al.
\newblock Activehne: Active heterogeneous network embedding.
\newblock In {\em Proc. of IJCAI}, 2019.

\bibitem[\protect\citeauthoryear{Cohn \bgroup \em et al.\egroup
  }{1996}]{cohn1996active}
David~A. Cohn, Zoubin Ghahramani, and Michael~I. Jordan.
\newblock Active learning with statistical models.
\newblock {\em J. Artif. Int. Res.}, 4(1):129?145, 1996.

\bibitem[\protect\citeauthoryear{Efron and Hinkley}{1978}]{efron1978assessing}
Bradley Efron and David~V Hinkley.
\newblock Assessing the accuracy of the maximum likelihood estimator: Observed
  versus expected fisher information.
\newblock {\em Biometrika}, 65(3):457--483, 1978.

\bibitem[\protect\citeauthoryear{Evans \bgroup \em et al.\egroup
  }{2014}]{Evans2014}
Craig Evans, Josh Friedman, Efe Karakus, and Jatin Pandey.
\newblock Potterverse.
\newblock \url{https://github.com/efekarakus/potter-network}, 2014.

\bibitem[\protect\citeauthoryear{Grover and
  Leskovec}{2016}]{grover2016node2vec}
Aditya Grover and Jure Leskovec.
\newblock node2vec: Scalable feature learning for networks.
\newblock In {\em Proc. of KDD}, pages 855--864, 2016.

\bibitem[\protect\citeauthoryear{Guillory and Bilmes}{2009}]{guillory2009label}
Andrew Guillory and Jeff~A Bilmes.
\newblock Label selection on graphs.
\newblock In {\em Proc. of NeurIPS}, pages 691--699, 2009.

\bibitem[\protect\citeauthoryear{Handcock \bgroup \em et al.\egroup
  }{2003}]{handcock2003statnet}
Mark~S Handcock, David Hunter, et~al.
\newblock Statnet: An {R} package for the statistical modeling of social
  networks.
\newblock {\em Web page http://www.csde.washington.edu/statnet}, 2003.

\bibitem[\protect\citeauthoryear{Jia \bgroup \em et al.\egroup
  }{2019}]{jia2019graph}
Junteng Jia, Michael~T Schaub, Santiago Segarra, and Austin~R Benson.
\newblock Graph-based semi-supervised \& active learning for edge flows.
\newblock In {\em Proc. of KDD}, pages 761--771, 2019.

\bibitem[\protect\citeauthoryear{Kang \bgroup \em et al.\egroup
  }{2019}]{kang2018conditional}
Bo~Kang, Jefrey Lijffijt, and Tijl {De Bie}.
\newblock Conditional network embeddings.
\newblock In {\em Proc. of ICLR}, 2019.

\bibitem[\protect\citeauthoryear{Kashima \bgroup \em et al.\egroup
  }{2009}]{kashima2009link}
Hisashi Kashima, Tsuyoshi Kato, et~al.
\newblock Link propagation: A fast semi-supervised learning algorithm for link
  prediction.
\newblock In {\em Proc. of SDM}, pages 1100--1111, 2009.

\bibitem[\protect\citeauthoryear{Kipf and Welling}{2017}]{kipf2016semi}
Thomas~N. Kipf and Max Welling.
\newblock Semi-supervised classification with graph convolutional networks.
\newblock In {\em Proc. of ICLR}, 2017.

\bibitem[\protect\citeauthoryear{Kong \bgroup \em et al.\egroup
  }{2011}]{kong2011dual}
Xiangnan Kong, Wei Fan, and Philip~S Yu.
\newblock Dual active feature and sample selection for graph classification.
\newblock In {\em Proc. of KDD}, pages 654--662, 2011.

\bibitem[\protect\citeauthoryear{Lehmann and Casella}{2006}]{lehmann2006theory}
Erich~L Lehmann and George Casella.
\newblock {\em Theory of point estimation}.
\newblock Springer Science \& Business Media, 2006.

\bibitem[\protect\citeauthoryear{Mart{\'\i}nez \bgroup \em et al.\egroup
  }{2017}]{martinez2017survey}
V{\'\i}ctor Mart{\'\i}nez, Fernando Berzal, and Juan-Carlos Cubero.
\newblock A survey of link prediction in complex networks.
\newblock {\em ACM Comput. Surv.}, 49(4):69, 2017.

\bibitem[\protect\citeauthoryear{Ostapuk \bgroup \em et al.\egroup
  }{2019}]{ostapuk2019activelink}
Natalia Ostapuk, Jie Yang, and Philippe Cudr{\'e}-Mauroux.
\newblock Activelink: deep active learning for link prediction in knowledge
  graphs.
\newblock In {\em Proc. of WWW}, pages 1398--1408, 2019.

\bibitem[\protect\citeauthoryear{Perozzi \bgroup \em et al.\egroup
  }{2014}]{perozzi2014deepwalk}
Bryan Perozzi, Rami Al-Rfou, and Steven Skiena.
\newblock Deepwalk: Online learning of social representations.
\newblock In {\em Proc. of KDD}, pages 701--710, 2014.

\bibitem[\protect\citeauthoryear{Rao}{1992}]{rao1992information}
C~Radhakrishna Rao.
\newblock Information and the accuracy attainable in the estimation of
  statistical parameters.
\newblock In {\em Breakthroughs in statistics}, pages 235--247. 1992.

\bibitem[\protect\citeauthoryear{Settles}{2009}]{settles2009active}
Burr Settles.
\newblock Active learning literature survey.
\newblock Technical report, University of Wisconsin-Madison Department of
  Computer Sciences, 2009.

\bibitem[\protect\citeauthoryear{Watts and
  Strogatz}{1998}]{watts1998collective}
Duncan~J Watts and Steven~H Strogatz.
\newblock Collective dynamics of 'small-world' networks.
\newblock {\em nature}, 393(6684):440, 1998.

\bibitem[\protect\citeauthoryear{Yang \bgroup \em et al.\egroup
  }{2014}]{yang2014active}
Zhilin Yang, Jie Tang, and Yutao Zhang.
\newblock Active learning for streaming networked data.
\newblock In {\em Proc. of CIKM}, pages 1129--1138, 2014.

\bibitem[\protect\citeauthoryear{Yang \bgroup \em et al.\egroup
  }{2016}]{yang2016revisiting}
Zhilin Yang, William~W Cohen, and Ruslan Salakhutdinov.
\newblock Revisiting semi-supervised learning with graph embeddings.
\newblock In {\em Proc. of ICML}, 2016.

\bibitem[\protect\citeauthoryear{Zafarani and Liu}{2009}]{Zafarani+Liu:2009}
R.~Zafarani and H.~Liu.
\newblock Social computing data repository at {ASU}, 2009.

\bibitem[\protect\citeauthoryear{Zhang and Oles}{2000}]{Zhang:Oles:00}
Tong Zhang and Frank~J. Oles.
\newblock A probability analysis on the value of unlabeled data for
  classification problems.
\newblock In {\em Proc. of ICML}, 2000.

\end{thebibliography}

\end{document}